\documentclass{article}
\usepackage{ijcai18}

\usepackage{times}
\usepackage{makecell}
\usepackage[utf8]{inputenc}
\usepackage[small]{caption}
\usepackage{graphicx}
\usepackage{subfig}
\usepackage{booktabs}
\usepackage{array}
\usepackage{float}
\usepackage{color}
\usepackage{natbib}
\usepackage{url}

\title{Vortex Pooling: Improving Context Representation in Semantic Segmentation}

\author{Chen-Wei Xie \quad Hong-Yu Zhou \quad Jianxin Wu\\
	National Key Laboratory for Novel Software Technology\\
	Nanjing University, China\\
	{\tt\small xiecw.mail@gmail.com, whuzhouhongyu@gmail.com, wujx2001@nju.edu.cn}
}

\begin{document}

\maketitle

\begin{abstract}

Semantic segmentation is a fundamental task in computer vision, which can be considered as a per-pixel classification problem. Recently, although \emph{fully convolutional neural network} (FCN) based approaches have made remarkable progress in such task, aggregating local and contextual information in convolutional feature maps is still a challenging problem. In this paper, we argue that, when predicting the category of a given pixel, the regions close to the target are more important than those far from it. To tackle this problem, we then propose an effective yet efficient approach named \emph{Vortex Pooling} to effectively utilize contextual information. Empirical studies are also provided to validate the effectiveness of the proposed method. To be specific, our approach outperforms the previous state-of-the-art model named DeepLab v3 by 1.5\% on the PASCAL VOC 2012 \emph{val} set and 0.6\% on the \emph{test} set by replacing the Atrous Spatial Pyramid Pooling (ASPP) module in DeepLab v3 with the proposed Vortex Pooling. Moreover, our model (10.13FPS) shares similar computation cost with DeepLab v3 (10.37 FPS).
\end{abstract}

\section{Introduction}

Nowadays, thanks to their powerful feature representations, deep convolutional neural network based approaches have achieved remarkable advances in various computer vision tasks, such as image classification, object detection and semantic segmentation. The goal of semantic segmentation is to assign a semantic label for each pixel in the given image. Most, if not all, of the state-of-the-art semantic segmentation models are base on the \emph{fully convolutional network} (FCN)~\citep{fcn}, in which all fully connected layers in models pre-trained on ImageNet are replaced by convolutional operations. As a result, FCN can take an arbitrary sized image as input and output a corresponding probability map, describing the probabilities of each pixel belonging to different semantic categories. However, due to the usage of strided pooling and convolution layers, the size of the probability map is usually much smaller than that of the original image, which makes it difficult to assign labels for \emph{every} pixel in the input image. To solve this issue,~\cite{fcn} tried using bilinear interpolation to upsample the probability map to the size of the input image.

The parameters of FCN are usually initialized from ImageNet pre-trained models and most of them are trained with images of size 224$\times$224, which are much smaller than the images in semantic segmentation datasets. For example, in the PASCAL VOC 2012 dataset~\citep{voc}, a large portion of images are about 500$\times$300 while the images in the Cityscapes dataset are usually about 2048$\times$1024. On the other hand, the receptive field of commonly used pre-trained models are not large enough. One typical instance is FCN-32s~\citep{fcn}, which is based on VGG16~\citep{vgg2014}, only has a theoretical receptive field of 404$\times$404. Moreover,~\cite{zhouboleirf} showed that the \emph{effective} receptive field is much smaller than the theoretical threshold, which may make the FCN only see a small part of the entire image.

\begin{figure}[t]
\centering
\includegraphics[height=4cm]{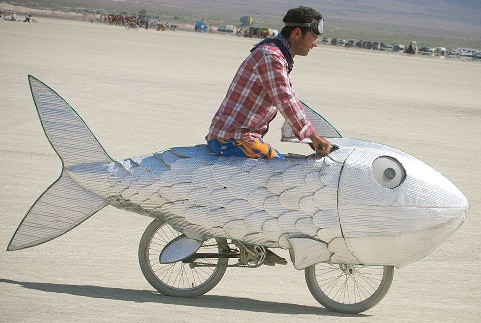}
\caption{In this image, if we only see the center part of the bike, the bike could be mistaken for a \emph{fish}. However, if we can have a glance at the whole image, it is easy to tell the object is a bike depending on its two obvious wheels and the person on the top of it. The original image can be found here.\protect\footnotemark[1]}
\label{fig:1}
\end{figure}
\footnotetext[1]{\url{https://github.com/BVLC/caffe/blob/master/examples/images/fish-bike.jpg}}

As for our human-beings, the contextual information is fairly important for us to classify pixels in images. As shown in Figure~\ref{fig:1}, when only seeing a part of the bike, even a person will (wrongly) think the object is a fish. But, if we check the entire image carefully, we could recognize that it is a bike because it has two wheels and a person is riding it. That is, a successful semantic segmentation method must have both large receptive field and be able to see the \emph{context}.

Recently, many approaches~\citep{lookwider,deeplabv3,pspnet,largekernelmatters} focused on aggregating the contextual information in semantic segmentation models. They have proven that context modules can significantly boost the performance. Specifically, their approaches usually have two stages: descriptor extraction and feature aggregation. The first stage aims at extracting convolutional feature maps for a given image. These feature maps, usually $H\times W\times C$ in size, can be regarded as $HW$ descriptors, each being $C$-dimension. These descriptors are good at capturing local details but lack of a global view. In the feature aggregation stage, these models often use different pooling strategies to aggregate the local and contextual information and generate another set of descriptors. Finally, each descriptor has local details as well as global contextual information.

DeepLab v3~\citep{deeplabv3} is one of the state-of-the-art semantic segmentation models. It uses ResNet-101~\citep{resnet} and their proposed ASPP module for descriptor extraction and feature aggregation, respectively. However, as will be discussed in our paper, the ASPP module \emph{only use a small part of all descriptors to classify each pixel}, which will lose some important contextual information.

In this paper, we aim at improving the feature aggregation approaches. The motivations can be summarized as follow: On one hand, when aggregating local and contextual information, we should \emph{consider as many descriptors as possible}, which helps us to get comprehensive context features for pixel-wise classification. On the other hand, although more descriptors are required, we should \emph{give different attention to them} because descriptors near the target pixel usually contain more related semantic information, and we are supposed to build a fine representation for it. For those descriptors far from the given pixel, a coarse representation should be enough.

According to these motivations, we propose \emph{Vortex Pooling}, a new context module that achieves both goals simultaneously. By \emph{replacing the ASPP module in DeepLab v3 with the proposed Vortex Pooling}, our semantic segmentation approach is able to achieve 84.2\% mean IoU on the PASCAL VOC 2012 \emph{val} set and 86.3\% mean IoU on the PASCAL VOC 2012 \emph{test} dataset, outperforming the state-of-the-art method DeepLab v3 by 1.5\% and 0.6\%, respectively.

\begin{figure}[t]
\centering
\includegraphics[height=2.82cm]{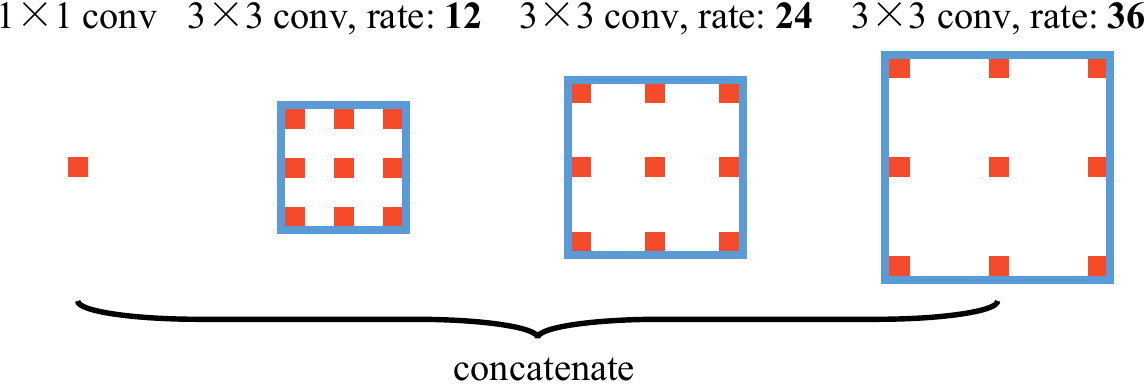}
\caption{Atrous Spatial Pyramid Pooling (ASPP).}
\label{aspp}
\end{figure}


\begin{figure*}[t]
    \centering
    \subfloat[]{\label{modela}
    \includegraphics[height=3.5cm]{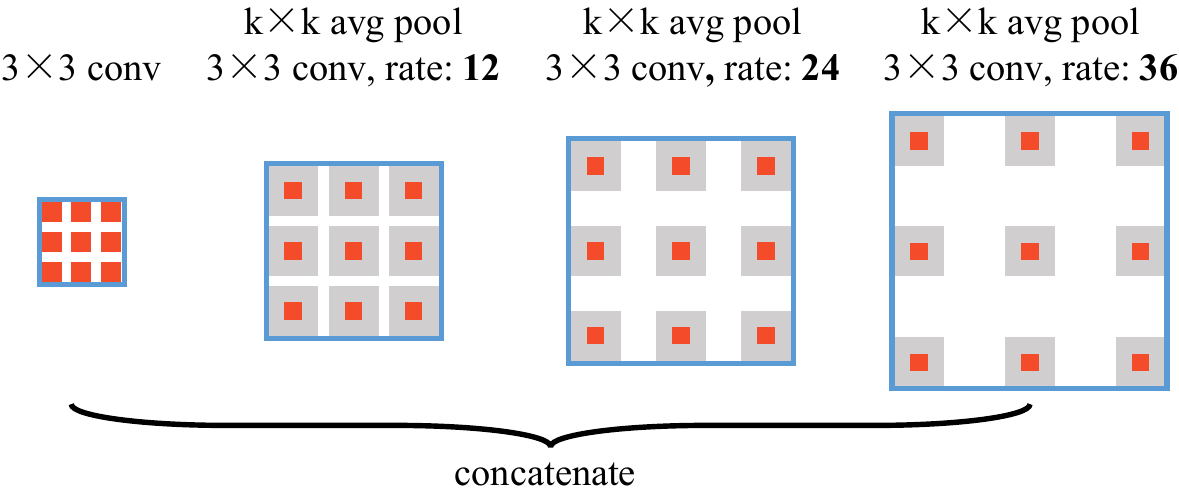}}
    \quad
    \qquad
    \subfloat[]{\label{modelb}
    \includegraphics[height=3.5cm]{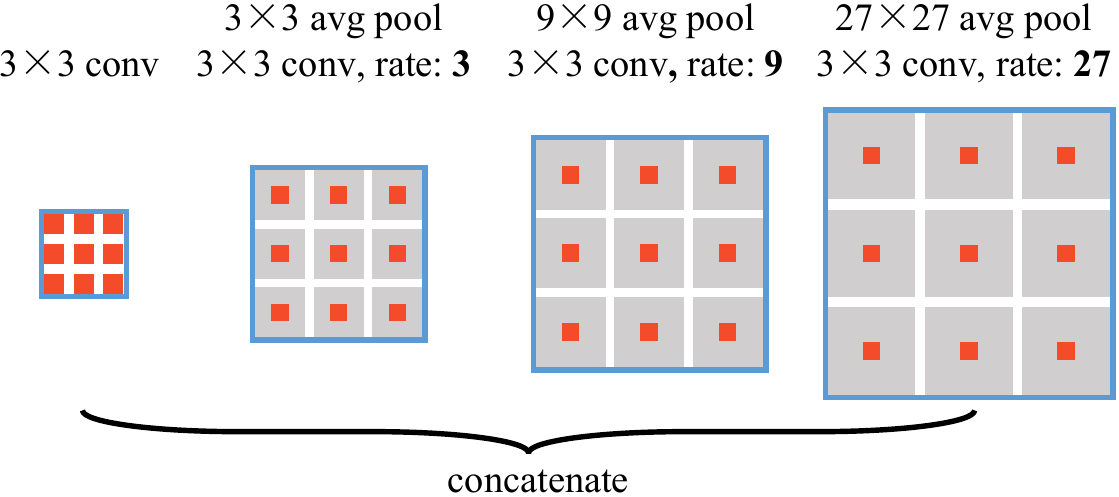}}\\

    \caption{(a), (b) show Module A and Module B, respectively. For the input feature map $X$, which is in size of $h\times w\times c_{1}$, both of these two module output $h\times w\times c_{2}$ tensor as new feature representations.}
    \label{Fig:RecAccuracy}
\end{figure*}
\section{Related Work}

Deep convolutional neural networks have made remarkable advances in image classification~\citep{cnnhinton}, object detection~\citep{fasterrcnn} and semantic segmentation~\citep{fcn,deeplabv1,deeplabv3}. For the semantic segmentation task,~\cite{fcn} proposed \emph{fully convolutional networks} (FCN) to take arbitrary sized image as input and output a 32$\times$ downsampled prediction. Note that in order to get a satisfactory probability map, they employed bilinear interpolation to upsample the prediction to the same size as its input. However, the downsampled feature maps may still lose too much detailed information, such as edges and texture, which is the reason why~\cite{fcn} tried to fuse feature maps from different intermediate layers to refine the final prediction. Unfortunately, features from intermediate layers usually have less semantic information compared with high level features and thus make the improvements limited.

The bilinear upsampling strategy used by~\cite{fcn} can be considered as a deconvolution operation with pre-defined parameters.~\cite{deconvnet} built an end-to-end system in which the parameters of the deconvolution layers can be learned from training data. Instead of using deconvolution layers,~\cite{tusimple_duc_multigrid} proposed Dense Upsamping Convolution (DUC) which directly predicted the full-resolution probability map by convolution.

To alleviate the low resolution feature map problem mentioned above,~\cite{dilation} proposed dilated convolution, which can increase the resolution of CNN feature maps and dramatically improve the accuracy of semantic segmentation models. However,~\cite{tusimple_duc_multigrid} and~\cite{dilatedresnet} showed that dilated convolution may cause ``gridding'' problems, and they proposed Hybrid Dilated Convolution (HDC) to remove such abnormal artifacts.

Some previous works focused on introducing prior information into semantic segmentation.~\cite{deeplabv1} used the \emph{conditional random field} (CRF) to model the compatibility between the predicted labels.~\cite{ial} argued that distinct classes should be of different importance for safe-driving (e.g., pedestrians are more important than sky in autonomous driving system).

Recently, many approaches confirmed that we can improve the performance of semantic segmentation models by incorporating contextual information.~\cite{lookwider} showed that a simple global average pooling feature can significantly improve the accuracy. DeepMask~\citep{deepmask} predicted the mask by adding a fully connected layer to utilize the global information. PSPNet~\citep{pspnet} used a pyramid pooling module to aggregate contextual information.~\cite{largekernelmatters} proposed separable convolution to approximate large convolution kernels so as to enlarge the receptive field.~\cite{dml} introduced a multi-label classification problem for the region near the target pixel.

DeepLab v3~\citep{deeplabv3} is one of the most recent state-of-the-art semantic segmentation models on multiple benchmarks. In their approach, they improved the ASPP module proposed in~\citep{deeplabv2} for better context features. The new ASPP module is composed by one 1$\times$1 convolution and three 3$\times$3 convolution with different dilation rates. In this paper, we delve into the ASPP module and explore its deficiency. Based on our discovery, we proposed \emph{Vortex Pooling}, which can aggregate features around the target position more efficiently by assigning different attention. By replacing ASPP with our proposed Vortex Pooling, our model outperforms DeepLab v3 by 1.5\% and 0.6\% mean IoU on the PASCAL VOC 2012 \emph{val} set and the \emph{test} set, respectively.

\section{The Proposed Method}

\subsection{DeepLab v3 recap}
Before we talk about the proposed method, we need to take a detour and briefly review DeepLab v3 and its ASPP module.

DeepLab v3 is a recent state-of-the-art approach in semantic segmentation. Given a $H\times W\times 3$ color image $I$ as input, DeepLab v3 feeds it to the feature net (e.g., ResNet-101) to get the feature map $X$, which is the output of the last convolution layer. $X$ is in size of $h\times w\times c_{1}$. For the best model in \citep{deeplabv3}, $H$ and $W$ are about 8 times larger than $h$ and $w$, respectively.

As aforementioned, each $c_{1}$-dimension descriptor in $X$ is lack of contextual information, so DeepLab v3 applies the Atrous Spatial Pyramid Pooling (ASPP) module on $X$ and gets new feature maps $Y_{aspp}$, which is in size of $h\times w\times c_{2}$. To incorporate global information, DeepLab v3 also applies global average pooling on $X$, feeds the result to a $1\times 1$ convolution with 256 filters, and then bilinearly upsample the feature to $h\times w$ in spatial to get image-level feature, which is denote as $Y_{g}$. $Y_{aspp}$ and $Y_{g}$ are concatenated and feeded to another $1\times1$ convolution layer and bilinear upsampling layer to get the final prediction.

The ASPP module described above is displayed in Figure~\ref{aspp}. It is composed by a $1\times 1$ convolution and three $3\times 3$ convolutions with dilation rates equal to (12, 24, 36), respectively. The output of these four convolution layer are \emph{concatenated} to get new feature maps, that is $Y_{aspp}$ aforementioned.

\subsection{Discussions and analyses} ASPP aggregates contextual information by multi-branch convolution with different dilation rates. Different dilation rates can dramatically increase the receptive field, but they can only perceive part of the object. Concretely, the input tensor $X$ has $h\times w$ descriptors, each being a $c_{1}$-dimension vector. As shown in Figure~\ref{aspp}, ASPP uses 25 descriptors in $X$ when computing each new feature descriptor in $Y_{aspp}$.\footnote{The 1$\times$1 convolution layer uses 1 descriptor, and the three 3$\times$3 convolution layers use $(3\times3)\times3=27$ descriptors. Note that the center descriptors of the three 3$\times$3 convolution layer are duplicate with the one of the 1$\times$1 convolution layer, so the descriptors used in total is $(1+27)-3=25$.}  However, $X$ usually have much more descriptors (e.g., $h$ and $w$ are both 65 if the input image is in size of 513$\times$513). We define the \emph{utilization ratio} as follows:
\begin{eqnarray}
	r=\frac{u}{hw}
\end{eqnarray}
where $u$ is the number of descriptors we used to aggregate each new descriptor in $Y_{aspp}$. So the utilization ratio of ASPP is $ r=\frac{25}{65 \times 65}\approx0.0059 $, which means that ASPP uses only 0.59\% of all descriptors to get each descriptor in $Y_{aspp}$. It shows that ASPP may lose some important contextual information. Based on this discovery, we propose two context modules that have higher utilization ratio and describe them in the next two sections.

\subsection{Module A}
We first propose Module A. Instead of perceiving 25 descriptors in ASPP, our Module A has the ability to utilize 25 subregions of the entire feature map.

We first take each $k\times k$ square region in $X$ as our subregion. As shown in Figure~\ref{modela}, note that we just highlight the currently used 25 subregions. Then we need to pool the descriptors in each subregion to one new descriptor. There are many pooling methods, such as average pooling, max pooling and second-order pooling~\citep{bilinear}. For simplicity, we use average pooling in our implementation.

After applying pooling operation, we use four convolution layers with different dilation rates to aggregate the descriptors from the 25 subregions. Obviously, the utilization ratio of our Module A is approximately $k^2$ larger than ASPP.

\begin{figure*}[t]
\centering
\includegraphics[height=6cm]{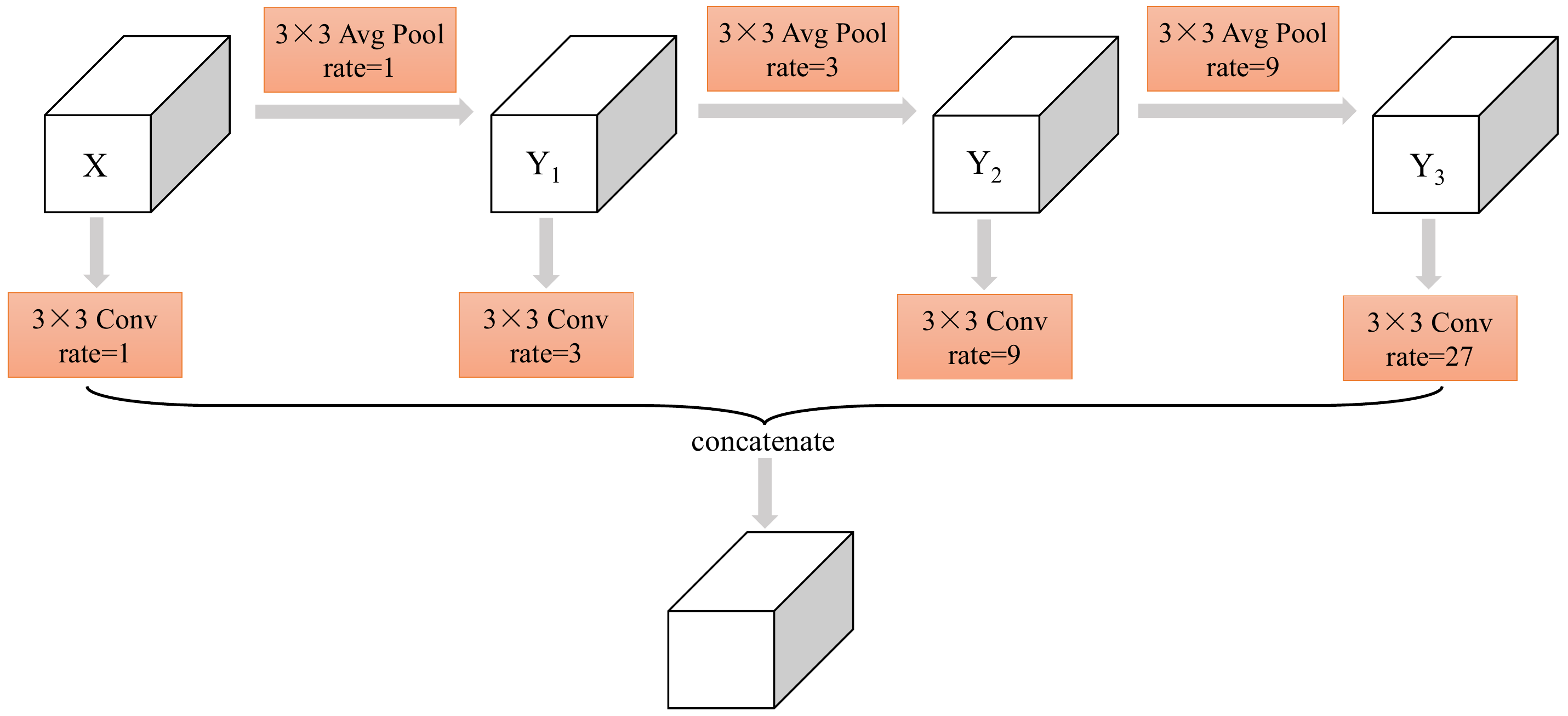}
\caption{An More efficient Vortex Pooling: Module C.}
\label{modelc}
\end{figure*}

\subsection{Module B}
We argue that the proposed Module A is still not that optimal because a typical issue of it is that, when aggregating sub-region features, Module A gives equal attention to all descriptors no matter whether they are near or far from the given pixel. However, as aforementioned, when classifying a pixel, the descriptors near from it may provide more related semantic information than that far way from it, so we need a fine representation for them. While the descriptors far way from it would rather provide contextual information instead of details, so we just use a coarse representation for them considering both the accuracy and efficiency.

In Figure~\ref{modelb}, we propose Module B, which is named \emph{Vortex Pooling}. To be specific, we use small $k$ for the subregions near from the given pixel, which enables more details. While for regions far away from the target pixel, we use large $k$ because only contextual information is needed. The $k$ values for the four convolution layers in Module B are set to (1, 3, 9, 27) respectively, which is a geometric sequence. Note that the geometric sequence can not only give different attention to different subregion, but also can be easily optimized for efficiently implementation, please refer to section~\ref{fast_vp} for more detail. Obviously, for the Module B displayed in Figure~\ref{modelb}, the utilization ratio is 1 if $h$ and $w$ are less than 81, which is easily held in practice.

\subsection{Accelerate Vortex Pooling}
\label{fast_vp}
The Vortex Pooling described above contains pooling operations with large kernel, which would be less efficient. So we proposed Module C to accelerate the proposed Vortex Pooling.

In the Module B, we apply three average pooling operations ($p_1$, $p_2$, $p_3$) with kernel size ($k=3$, $k^{2}=9$, $k^{3}=27$) on the same activation map $X$, and get the results $Y_1$, $Y_2$ and $Y_3$, as shown below (we ignore the normalization here for simplity).
\begin{eqnarray}
(Y_{1})_{i,j}=p_1(X)=\sum_{m=i-(k-1)/2}^{i+(k-1)/2}\sum_{n=j-(k-1)/2}^{j+(k-1)/2}X_{m,n}\\
(Y_{2})_{i,j}=p_2(X)=\sum_{m=i-(k^2-1)/2}^{i+(k^2-1)/2}\sum_{n=j-(k^2-1)/2}^{j+(k^2-1)/2}X_{m,n}\\
(Y_{3})_{i,j}=p_3(X)=\sum_{m=i-(k^3-1)/2}^{i+(k^3-1)/2}\sum_{n=j-(k^3-1)/2}^{j+(k^3-1)/2}X_{m,n}
\label{vp}
\end{eqnarray}
We note that we can reuse $Y_1$ while computing $Y_2$
\begin{eqnarray}
\centering
(Y_2)_{i,j}=\sum_{p=-(k-1)/2}^{(k-1)/2}\sum_{q=-(k-1)/2}^{(k-1)/2}(Y_1)_{i+pk,j+qk}
\label{vp_prove}
\end{eqnarray}
In practice, we can compute $Y_2$ efficiently by

\begin{eqnarray}
Y_{2}=p_{2}(X)=p_{2}'(p_{1}(X))=p_{2}'(Y_1)
\label{vp_faster}
\end{eqnarray}

where $p_2'$ is a $k\times k$ average pooling with dilation rate $k$. Note that we should multiply some additional coefficients to make equation~\ref{vp_prove} hold on the border of $X$.

We can also reuse $Y_2$ while we computing $Y_{3}$, as the Module C shown in Figure~\ref{modelc}. Although Module B and Module C are equivalent in mathematics, Module C is much more efficient than Module B. In our implementation, Module B use a $3\times3$ average pooling, a $9\times9$ average pooling and a $27\times27$ average pooling, while Module C only use three $3\times3$ average pooling. Experiments show that our segmentation model with Module C shares similar inference speed with DeepLab v3 but achieves higher accuracy.

\section{Experimental Results}

\begin{figure*}[t]
\centering
\includegraphics[height=7cm]{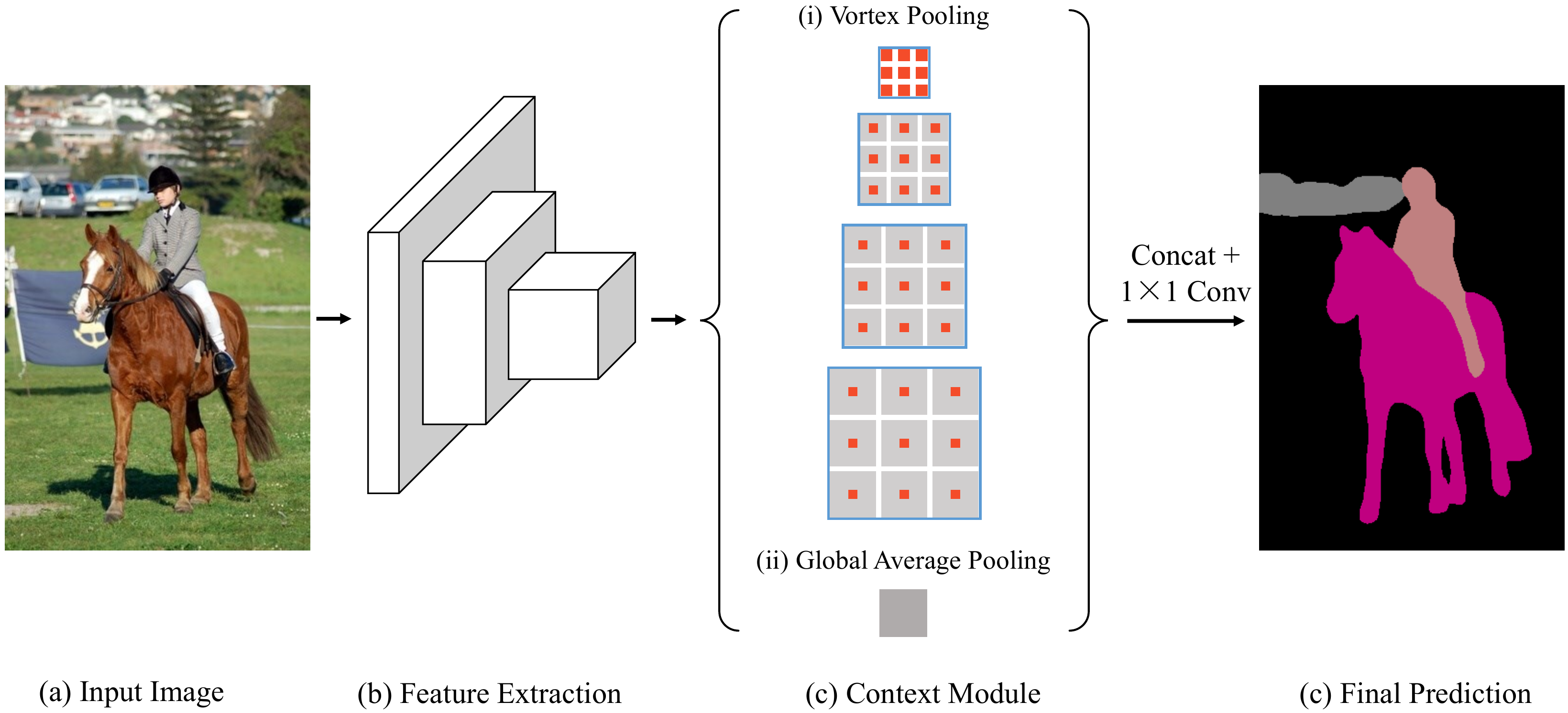}
\caption{Overview of our proposed semantic segmentation model. Given an image as input, we first extract the feature of the last convolutional neural network, then use our context module to get the multi-level context feature and concatenate them into one single descriptor. Following~\cite{deeplabv3}, we also use a global average pooling to incorporate global feature. At last, we use a $1\times 1$ convolution layer to get the final probability map.}
\label{framework}
\end{figure*}

In this section, we evaluate the proposed methods on the PASCAL VOC 2012 dataset and the Microsoft COCO dataset~\citep{coco}. We first use ResNet-50 as our basic model in the ablative experiments for its good trade-off between accuracy and computation complexity. Then we use our best context module based on ResNet-101 to compete with state-of-the-art approaches.

\subsection{Ablation Studies}
\label{abstudy}
We use the PASCAL VOC 2012 dataset for ablation studies. The official dataset has 1,464 training images, 1,449 validation images and 1,456 test images.~\cite{vocaug} augmented this dataset by providing extra pixel-wise annotations, resulting in 10,582 training images. We use these 10,582 images as the \emph{train\_aug} set for training following the instructions in~\citep{deeplabv2,deeplabv3} and use the official PASCAL VOC 2012 \emph{val} set for evaluation.

We keep the same hyper-parameters and training strategy as those in~\citep{deeplabv3}. Concretely, details of our training protocol are listed below:

\textbf{Learning rate policy:} Following~\cite{deeplabv3}, we use the ``poly" learning rate policy with initial learning rate 0.007, and the learning rate is multiplied by $(1-\frac{iter}{max\_iter})^{power}$ at each iteration, where $power$ is set to 0.9. We train the model for 30,000 iterations.

\textbf{Data Augmentation:} Similar to~\cite{deeplabv3}, we randomly scale the images from 0.5 to 2.0 and randomly flip the input images in a left-right manner during the training stage.

\textbf{Crop Size:}~\cite{deeplabv3} and~\cite{pspnet} showed that large crop size can significantly improve the accuracy, so that we crop a $513\times 513$ patch from every resized image for training following~\cite{deeplabv3}.

\textbf{Multi-grid:}~\cite{tusimple_duc_multigrid} and~\cite{dilatedresnet} demonstrated that the dilated convolution can lead to gridding artifacts. ~\cite{tusimple_duc_multigrid} and~\cite{deeplabv3} applied a multi-grid method for the ResNet to alleviate the gridding artifacts. For the three bottleneck blocks in \emph{block4} of ResNet, \cite{deeplabv3} multiplied the dilation rate by (1, 2, 4), respectively. For convenience, we follow the same setting as in~\citep{deeplabv3}.

\textbf{Inference strategy on the val set:} During the inference stage, the $output\_stride$ is set to 8 (dilated convolution are used in the last two blocks of ResNet) which is the same as the setting when training models. For all ablative experiments, we evaluate the model on single-scaled images without any left-right flipping.

\textbf{Batch Normalization:} We use $batch size = 16$ for all of our experiments. In~\citep{resnet,deeplabv2}, they fixed the parameters of batch normalization layer while fine-tuning ResNet, however,~\cite{deeplabv3} and~\cite{pspnet} showed that updating the batch normalization layer can significantly increase the accuracy of semantic segmentation models. It is worth noting that fine-tuning the batch normalization while training is non-trivial, because the multi-GPU implementation of batch normalization in popular deep learning frameworks (e.g. Caffe, Torch, Tensorflow, Pytorch) compute $mean$ and $variance$ within each \emph{single} GPU (the whole batch is distributed on different GPUs). It would not be a problem in some tasks when the batch size is very large, such as image classification. For semantic segmentation, the sub-batchsize on a single GPU is usually less than 5, which makes it hard to approximate the global expectation and variance computed on the whole batch. We then determine to use the synchronized batch nomalization layer implemented by~\cite{syncbn} to synchronize the statistics such as $mean$ and $variance$ across multiple GPUs.

The ablative results are showed in Table~\ref{table:1}. ASPP can obtain 75.4\% mean IoU on PASCAL VOC 2012 \emph{val} set. We also try replacing the $1\times 1$ convolution layer in ASPP to $3\times 3$ convolution layer, and denote it as ASPP+. The ASPP+ gains small improvement over ASPP.

We then replace ASPP to our proposed Module A. With $k=5$, the Module A ($5\times 5$) achieves better accuracy than the ASPP. However, increasing the pooling size cannot further improve the accuracy. For example, Module A ($9\times 9$) is even worse than Module A ($5\times 5$) although its $9\times 9$ pooling kernel obtains a higher utilization ratio (0.479) than $5\times 5$ does (0.148). We can tell that the utilization ratio is not the only factor that impacts the accuracy. As for the reason, we argue that for the region near the given pixel, $9\times 9$ average pooling is so large that lots of details are discarded (or eliminated), while for the region far from the given pixel, a $9\times 9$ kernel can be too small for comprehensive contextual pooling.

Our Module B can avoid the problem mentioned above. It utilizes average pooling with different kernel size. The pooling with small kernel size can be used to build fine feature representations of details, while the large one can help to get coarse contextual information. For the kernel size, we just use (1, 3, 9, 27) for simplicity. Our Module B achieves the best performance in all three approaches.

\begin{table}
\caption{Ablation results on the PASCAL VOC 2012 \emph{val} set. ASPP+ means replacing the 1$\times$1 convolution kernel of ASPP with 3$\times$3 kernel. Module A ($5\times 5$) means using $k=5$ in Module A.}
\centering
\begin{tabular}{c|ccc}
\toprule
Module & Mean IoU\\
\midrule
ASPP & 75.4\\
ASPP+ & 75.6\\
Module A ($5\times 5$) & 76.1\\
Module A ($9\times 9$) & 75.6\\
Module A ($13\times 13$) & 75.8\\
Module B & 76.6\\
\bottomrule
\end{tabular}
\label{table:1}
\end{table}

\subsection{Compared with State-of-the-Art}

In this section, we use our Vortex Pooling to compete with the previous state-of-the-art approaches. Following most previous works, we use ResNet-101 as our basic model. The framework is shown in Figure~\ref{framework}. For a given image, we extract the convolution feature maps, and use our Vortex Pooling for feature encoding, following~\cite{deeplabv3}, we also use a global average pooling to incorporate global feature. Then a $1\times 1$ convolution layer is applied to get the final prediction.

We start from the ResNet-101 pre-trained on ImageNet, then use the Microsoft COCO and the PASCAL VOC 2012 for fine-tuning. Note that some previous works used more data than us for training. The DeepLab v3-JFT in~\citep{deeplabv3} employed a more powerful basic model pre-trained on both ImageNet and JFT-300M (containing about 300 million images), so we do not compare with it for fair. ~\cite{idw} and~\cite{dis} used the \emph{Image Descriptions in the Wild} (IDW) dataset for semi-supervised learning, which is out of the range of this paper. We evaluate our model on the PASCAL VOC 2012 \emph{val} set and the \emph{test} set, respectively. The details of our experiments are listed as follow:

\begin{figure*}[!t]
\centering
\subfloat[\scriptsize{Image}]{
\begin{minipage}[b]{0.115\textwidth}
\includegraphics[width=1\textwidth]{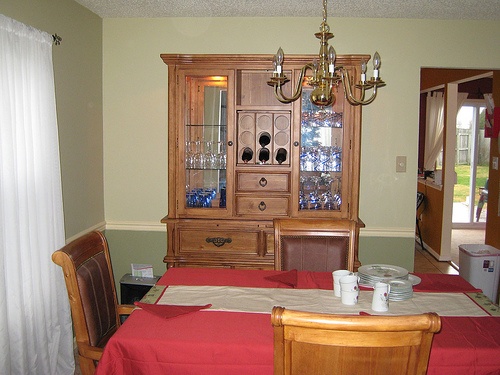}\\
\includegraphics[width=1\textwidth]{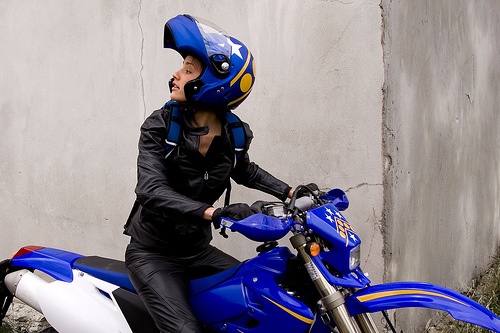}\\
\includegraphics[width=1\textwidth]{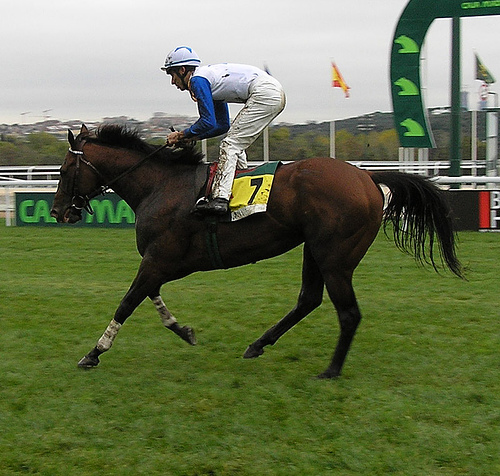}\\
\includegraphics[width=1\textwidth]{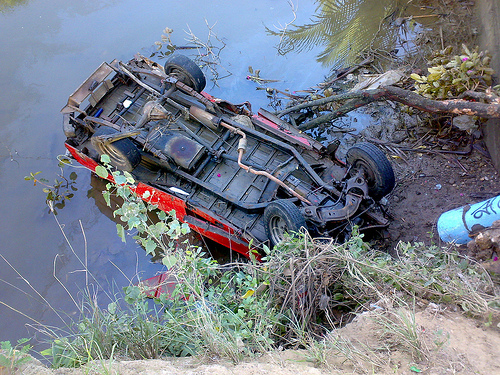}
\end{minipage}
}
\subfloat[\scriptsize{Ground Truth}]{
\begin{minipage}[b]{0.115\textwidth}
\includegraphics[width=1\textwidth]{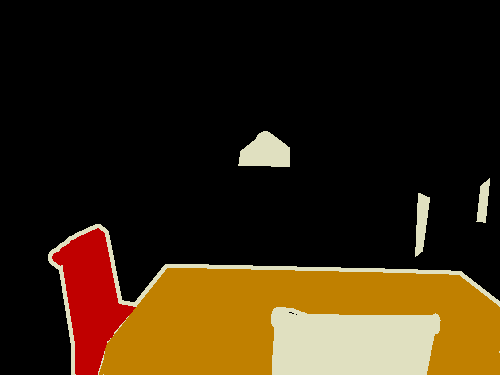}\\
\includegraphics[width=1\textwidth]{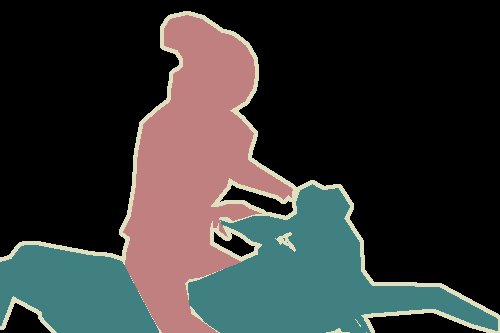}\\
\includegraphics[width=1\textwidth]{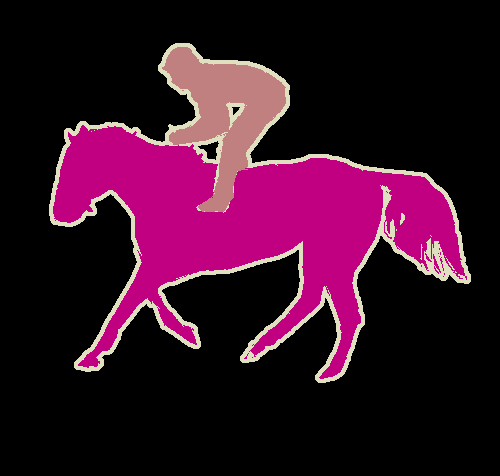}\\
\includegraphics[width=1\textwidth]{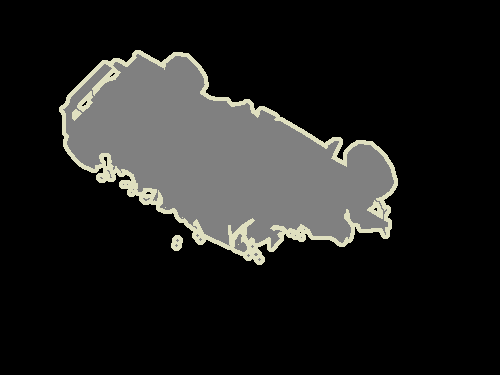}
\end{minipage}
}
\subfloat[\scriptsize{DeepLab v3}]{
\begin{minipage}[b]{0.115\textwidth}
\includegraphics[width=1\textwidth]{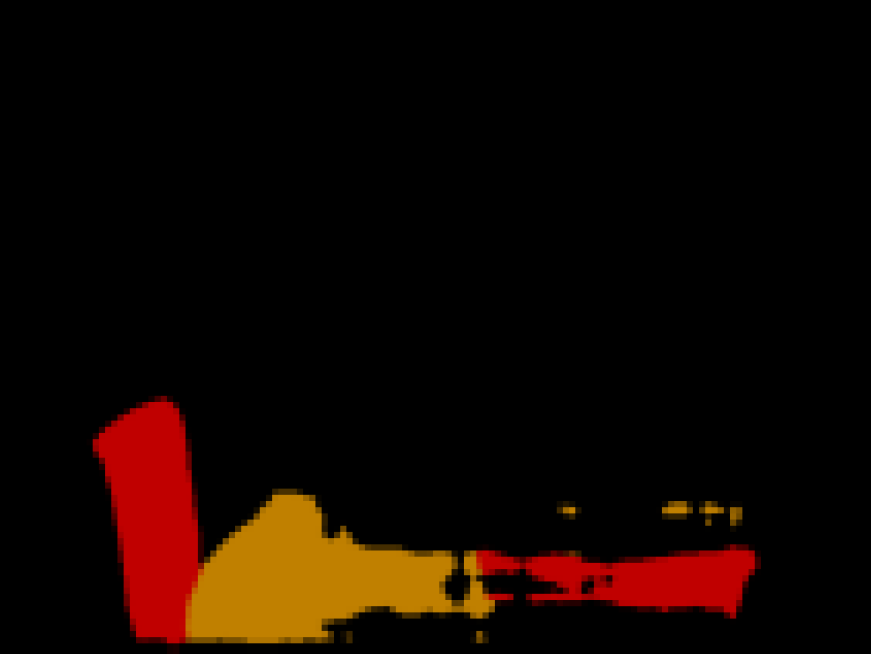}\\
\includegraphics[width=1\textwidth]{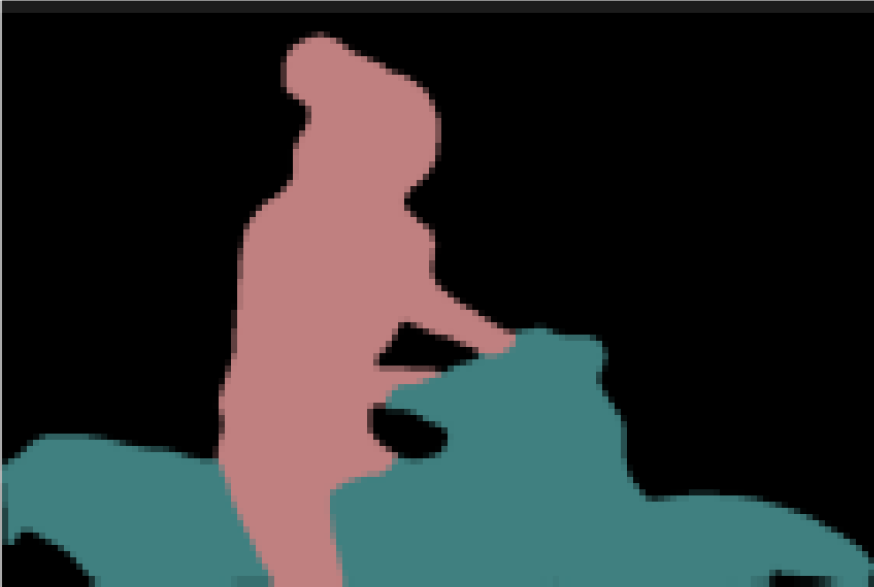}\\
\includegraphics[width=1\textwidth]{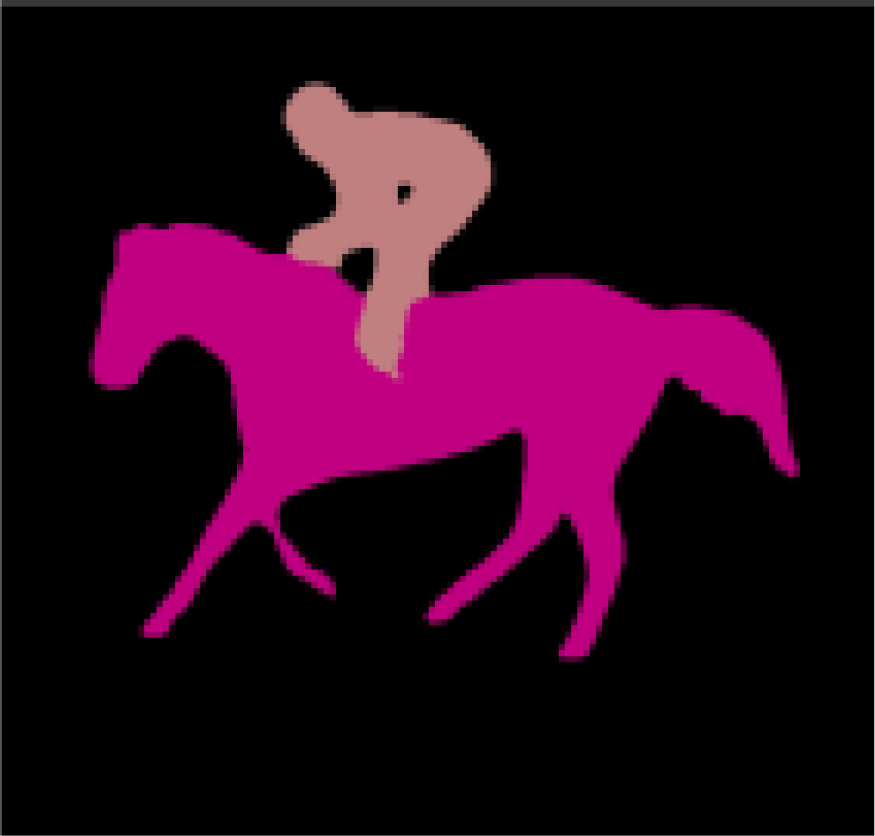}\\
\includegraphics[width=1\textwidth]{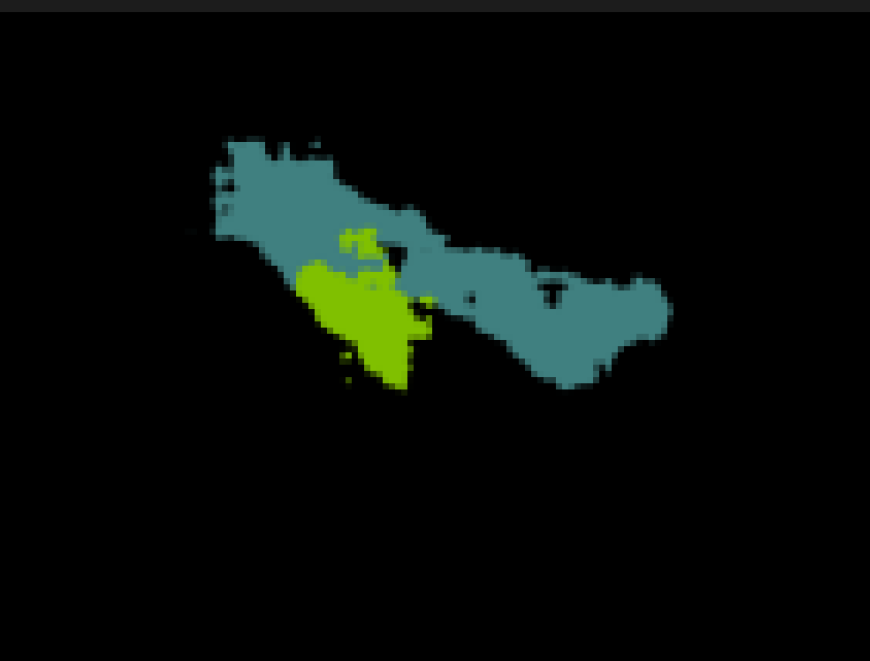}
\end{minipage}
}
\subfloat[\scriptsize{Ours}]{
\begin{minipage}[b]{0.115\textwidth}
\includegraphics[width=1\textwidth]{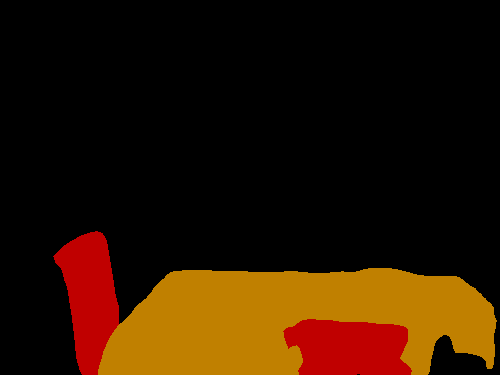}\\
\includegraphics[width=1\textwidth]{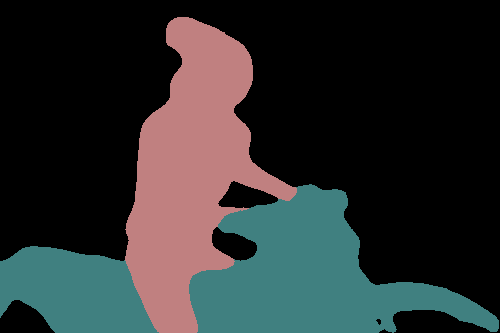}\\
\includegraphics[width=1\textwidth]{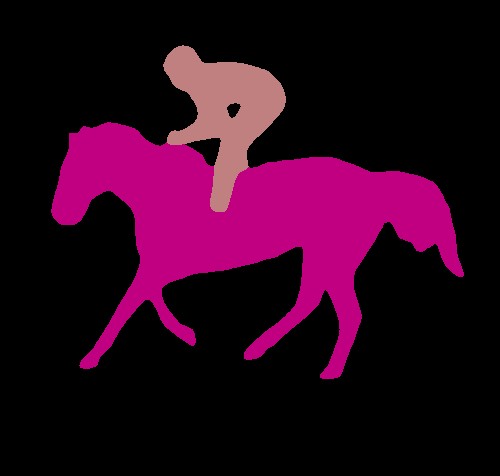}\\
\includegraphics[width=1\textwidth]{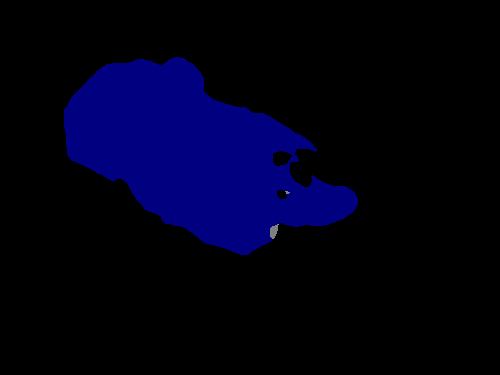}
\end{minipage}
}
\subfloat[\scriptsize{Image}]{
\begin{minipage}[b]{0.115\textwidth}
\includegraphics[width=1\textwidth]{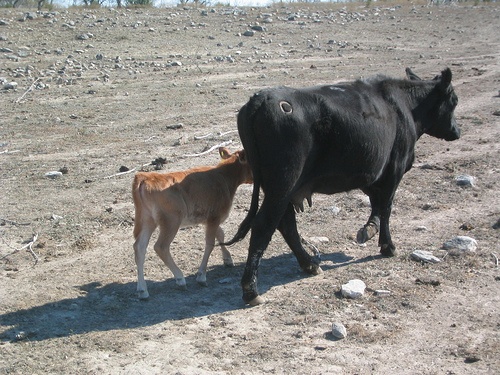}\\
\includegraphics[width=1\textwidth]{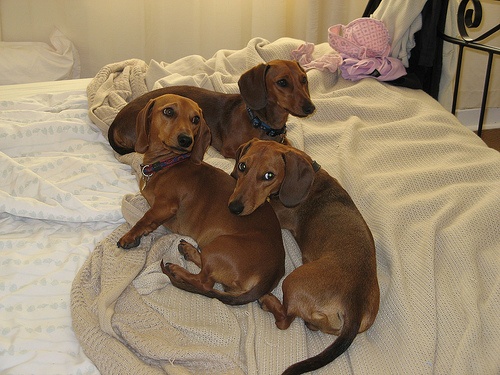}\\
\includegraphics[width=1\textwidth]{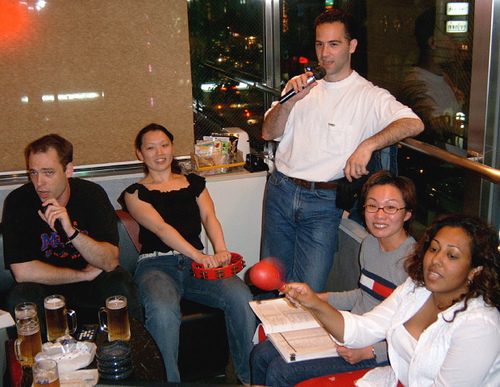}\\
\includegraphics[width=1\textwidth]{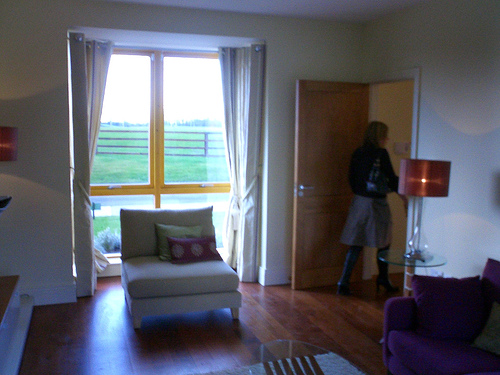}
\end{minipage}
}
\subfloat[\scriptsize{Ground Truth}]{
\begin{minipage}[b]{0.115\textwidth}
\includegraphics[width=1\textwidth]{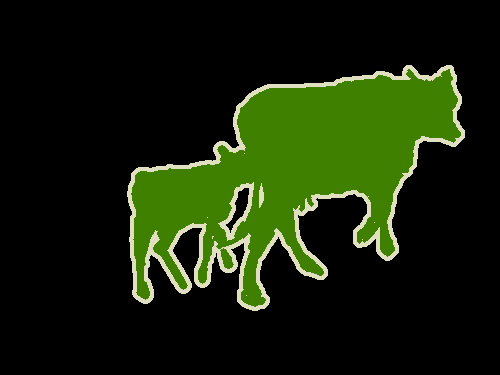}\\
\includegraphics[width=1\textwidth]{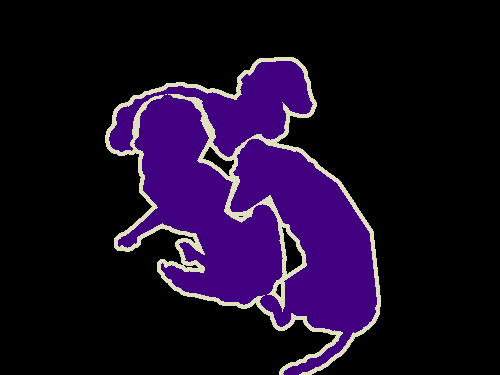}\\
\includegraphics[width=1\textwidth]{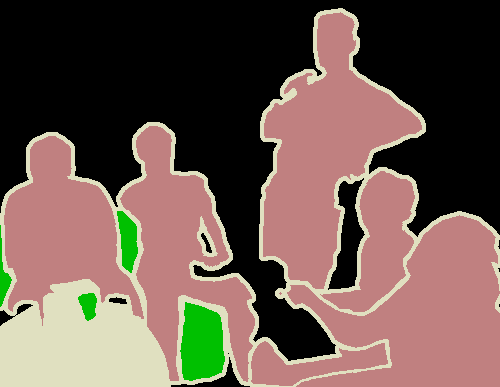}\\
\includegraphics[width=1\textwidth]{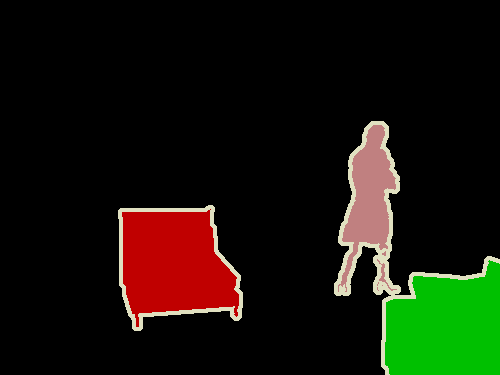}
\end{minipage}
}
\subfloat[\scriptsize{DeepLab v3}]{
\begin{minipage}[b]{0.115\textwidth}
\includegraphics[width=1\textwidth]{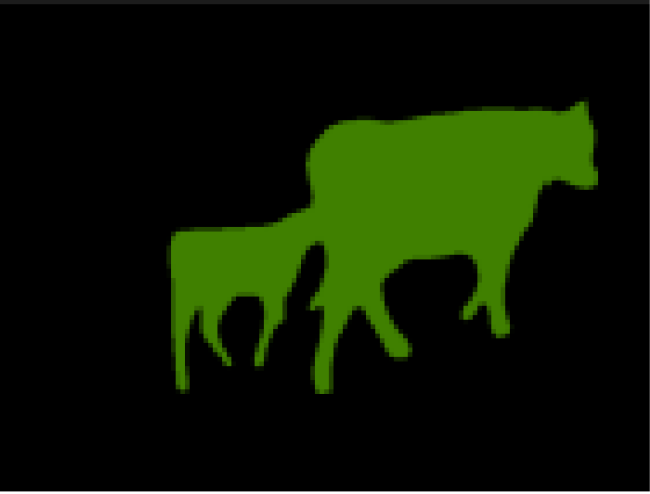}\\
\includegraphics[width=1\textwidth]{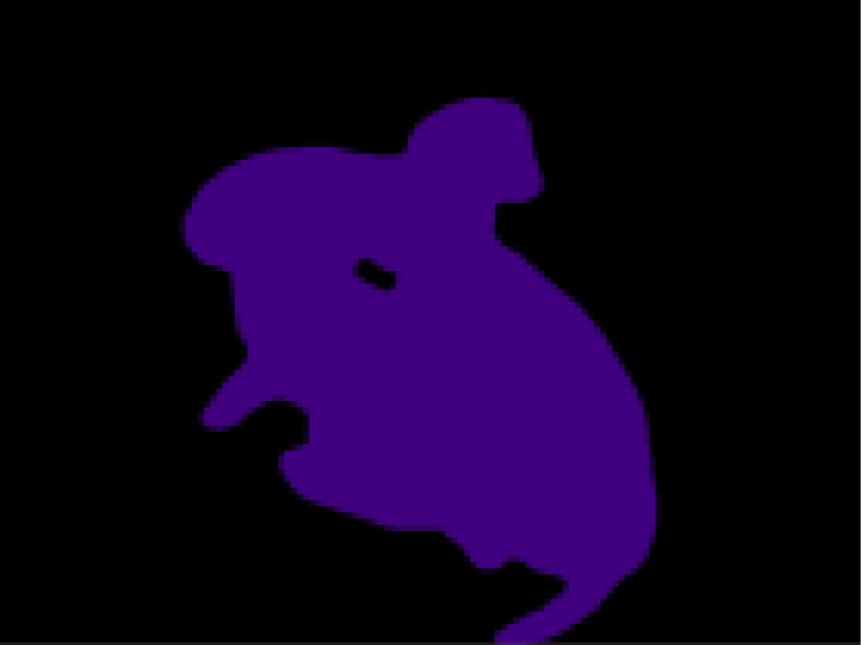}\\
\includegraphics[width=1\textwidth]{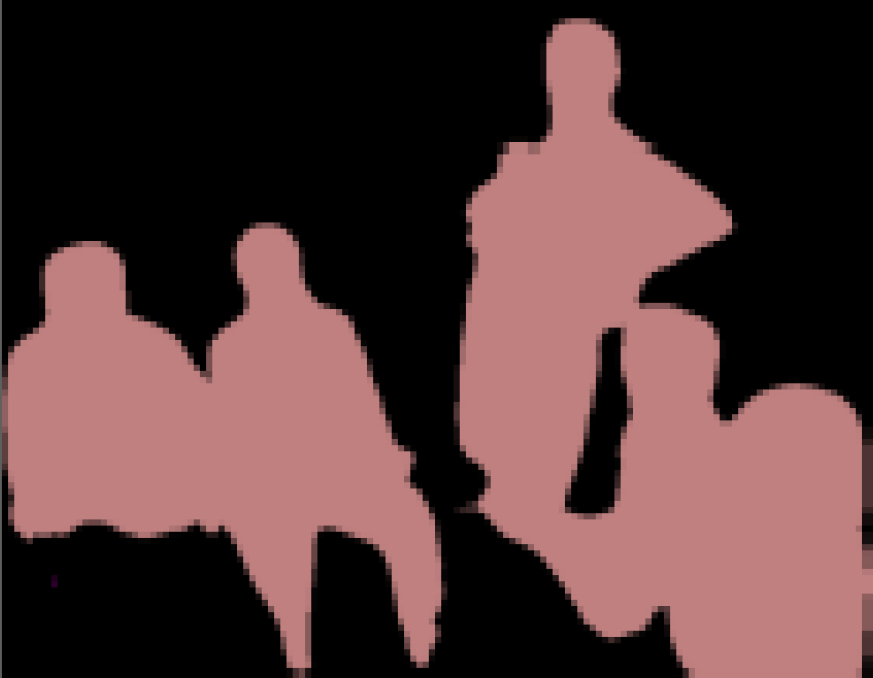}\\
\includegraphics[width=1\textwidth]{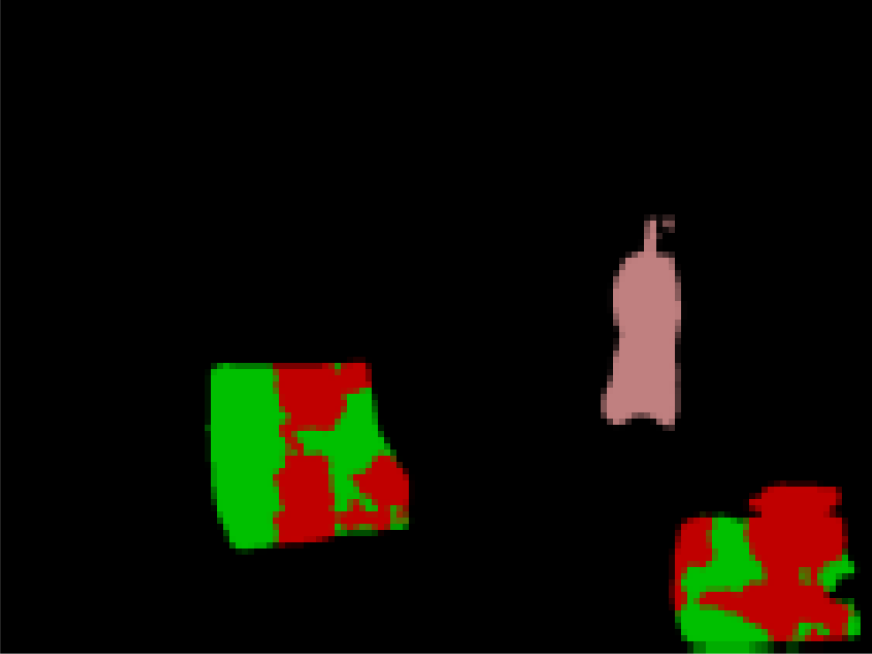}
\end{minipage}
}
\subfloat[\scriptsize{Ours}]{
\begin{minipage}[b]{0.115\textwidth}
\includegraphics[width=1\textwidth]{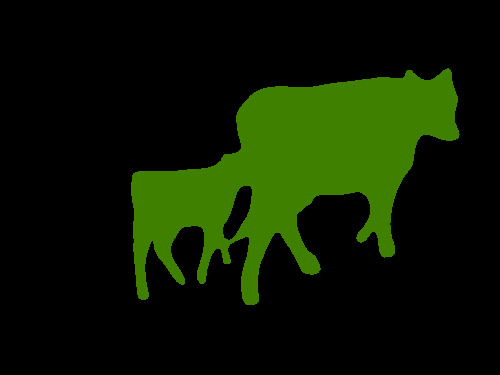}\\
\includegraphics[width=1\textwidth]{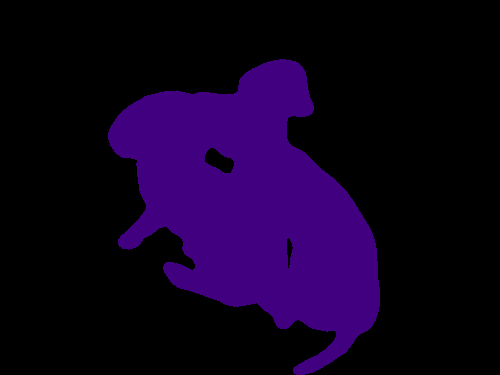}\\
\includegraphics[width=1\textwidth]{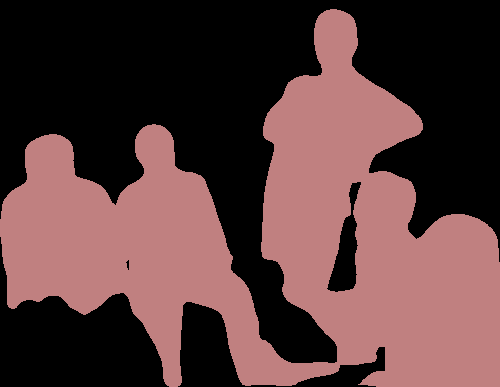}\\
\includegraphics[width=1\textwidth]{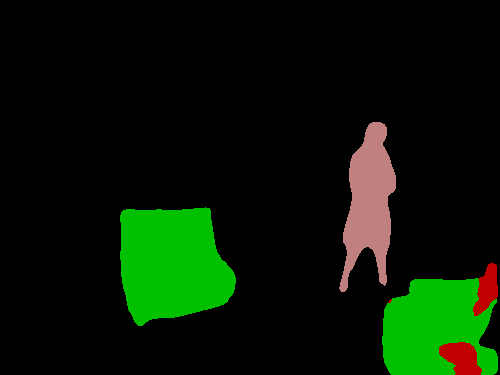}
\end{minipage}
}
\caption{Visualized comparison between DeepLab v3 and our approach on the PASCAL VOC 2012 \emph{val} set. Our model uses a more powerful context module so we can recognize some difficult pixel (e.g., the chair and table in the first image). Note that the last row shows two failure cases. For the failure case in the left,  the car is predicted to be a ship by our model, we conjecture that the model makes this decision because of the water around it. While for the failure case in the right, the chair is mistaken for a sofa but we can see that the difference between them can be limited. (This figure is best viewed in color.)}
\label{visualization}
\end{figure*}

\textbf{Pretrained On MS COCO:} The MS COCO dataset has 80 categories, which is larger than the number of classes in PASCAL VOC dataset. Following the steps in~\citep{deeplabv3}, we treat the categories which are not defined in PASCAL VOC 2012 as background and discard the objects which are smaller than 1000 pixels. We also use the same learning rate policy as that in~\citep{deeplabv3}. Specifically, we set the initial learning rate to 0.007 and train the model for 200,000 iterations.

\textbf{Fine-tuned on PASCAL VOC 2012:}
Although MS COCO provides more training images than PASCAL VOC, the former uses a polygon to approximate the mask of each object, which provides a lower quality of annotations. So~\cite{largekernelmatters} and~\cite{deeplabv3} employed a three-stage training strategy. To be specific,~\cite{deeplabv3} firstly trains the model with MS COCO only, and then they use PASCAL VOC 2012 augmented set for further training. Finally, they fine-tune the model on official PASCAL VOC 2012 data. Note that~\cite{deeplabv3} duplicated the images which contain hard categories (bicycle, chair, table, potted plant, and sofa) in the training set, but we find it is useless in our experiments, so we do \emph{not} use any hard example mining while training our model.

\textbf{Evaluation on the PASCAL VOC 2012 val set:}
With the MS COCO as additional training data, we first evaluate our model on the PASCAL VOC 2012 \emph{val} set. Most state-of-the-art approaches use multi-scale and left-right flipping tricks while test. For example,~\cite{deeplabv3} uses \emph{scale} = (0.5, 0.75, 1.0, 1.25, 1.5, 1.75) during the inference stage, and they can achieve 82.7\% mean IoU on the PASCAL VOC 2012 \emph{val} set. Following the same setting, our model can achieve 84.2\% mean IoU, 1.5\% higher than the DeepLab v3 counterpart. The detailed results are displayed in Table~\ref{12val}. We also provide some visualization in Figure~\ref{visualization}, including both successful examples and failure cases. Our model uses a more powerful context module so we can recognize some difficult pixel (e.g., the chair and table in the first image). For our failure cases, the image in the left predicts the \emph{car} to be a \emph{ship}, we conjecture that our model makes this decision according to the water around it. For the failure case in the right, our model outputs a \emph{chair} which looks very similar to a \emph{sofa} (see the sofa on the right side).

\begin{table}[!htbp]
\small
\centering
\caption{Performance on the PASCAL VOC 2012 \emph{val} set with the MS COCO and the PASCAL VOC 2012 dataset for training. Our model is 1.5\% higher than the DeepLab v3 counterpart.}
\begin{tabular}{p{6cm}|p{1.5cm}<{\centering}}
\toprule
Method & mIoU\\
\midrule
DeepLabv1-CRF~\citep{deeplabv1} & 68.7\\
Front + Large + RNN~\citep{dilation} & 73.9\\
DeepLabv2-CRF~\citep{deeplabv2} & 77.7\\
HikSeg COCO~\citep{hikseg} & 80.6\\
Large\_Kernel\_Matters~\citep{largekernelmatters} & 81.0\\
DeepLab v3~\citep{deeplabv3} & 82.7\\
\hline
Ours & \textbf{84.2} \\
\bottomrule
\end{tabular}
\label{12val}
\end{table}

\textbf{Inference on test set}
We also evaluate our model on the PASCAL VOC 2012 \emph{test} set. Following~\citep{deeplabv3}, we first use the official PASCAL VOC 2012 \emph{trainval} set for fine-tuning before evaluating on the \emph{test} set, and then we submit our results to the online evaluation server. Our model obtains 86.3\% mean IoU on PASCAL VOC 2012 \emph{test} set, and the anonymous submission link can be found here,\footnote{\url{http://host.robots.ox.ac.uk:8080/anonymous/KWFTE2.html}} please see Table~\ref{12test} for details. Note that we do \emph{not} use CRF for post-processing. Our model outperforms DeepLab v3 by 0.6\%.

\begin{table}[htbp]
\small
\centering
\caption{Performance on the PASCAL VOC 2012 \emph{test} set with the MS COCO and the PASCAL VOC 2012 dataset for training. Our model is 0.6\% higher than the DeepLab v3 counterpart.}
\begin{tabular}{p{6cm}|p{1.5cm}<{\centering}}
\toprule
Method & mIoU\\
\midrule
Piecewise~\citep{piecewise} & 78.0\\
DeepLabv2-CRF~\citep{deeplabv2} & 79.7\\
HikSeg COCO~\citep{hikseg} & 81.4\\
SegModel~\citep{segmodel} & 82.5\\
Layer Cascade~\citep{notallpixelareequal} & 82.7\\
TuSimple~\citep{tusimple_duc_multigrid} & 83.1\\
Large\_Kernel\_Matters~\citep{largekernelmatters} & 83.6\\
Multipath-RefineNet & 84.2\\
ResNet-38\_MS\_COCO~\citep{widerordeeper} & 84.9\\
PSPNet~\citep{pspnet} & 85.4\\
DeepLab v3~\citep{deeplabv3} & 85.7\\
\hline
Ours & \textbf{86.3} \\
\bottomrule
\end{tabular}
\label{12test}
\end{table}

\textbf{Inference speed}
We also compare the inference time of our model with DeepLab v3. Both of our Vortex Pooling and DeepLab v3 use ResNet-101 as backbone network and implemented in Pytorch. Given images with size equal to $513\time 513$ as input, on one NVIDIA TITAN Xp GPU, the inference time of different models are shown in Table~\ref{speed}. Note that due to the large kernel size pooling, the Module B is slower than DeepLab v3, while our Module C has the same accuracy with Module B but can be more efficient.
\begin{table}[htbp]
\small
\centering
\caption{Inference speed on $513\times 513$ image.}
\begin{tabular}{p{6cm}|p{1.5cm}<{\centering}}
\toprule
Method & FPS\\
\midrule
DeepLab v3~\citep{deeplabv3} & 10.37\\
\hline
Our Module B & 8.06 \\
Our Module C & 10.13 \\
\bottomrule
\end{tabular}
\label{speed}
\end{table}


\section{Conclusion}
In this paper, we explored the drawback of the state-of-the-art approach DeepLab v3 in semantic segmentation, then proposed a new context module based on two discoveries. Firstly, when classifying a pixel in an image, we should consider as many descriptors as possible to get comprehensive contextual information. Secondly, the descriptors near this pixel are more important than those far from it. The descriptors near the target pixel usually contain more related semantic information, so we use average pooling with small kernel to get fine representations. While the descriptors far from the target pixel mainly provide contextual information, so coarse representations are enough. Experimental results show the effectiveness of our proposed module.

\bibliographystyle{named}
\bibliography{ijcai18}

\end{document}